\documentclass[a4paper, reqno]{article}
\usepackage{latexsym}
\usepackage{graphicx}
\usepackage{color}
\usepackage[english]{babel}
\usepackage[utf8x]{inputenc}
\usepackage[T1]{fontenc}
\usepackage{cleveref}
\usepackage[euler]{textgreek}
\usepackage{float}
\usepackage[a4paper,top=3cm,bottom=2cm,left=3cm,right=3cm,marginparwidth=1.75cm]{geometry}
\usepackage{amsmath}
\usepackage{verbatim}
 
\usepackage{graphicx}
\usepackage[colorinlistoftodos]{todonotes}
\usepackage{amssymb}
\usepackage{amsmath}
\usepackage{txfonts}
\usepackage{mathdots}
\usepackage[classicReIm]{kpfonts}

\begin{document}

\begin{center}
{\Large\textbf{Image denoising and restoration with CNN-LSTM Encoder Decoder 
with Direct Attention}}
\vspace{4\baselineskip}

$^{\textcolor{blue}{1}}$\textcolor{blue}{Kazi Nazmul 
Haque, 
}
$^{\textcolor{blue}{2}}$\textcolor{blue}{Mohammad 
Abu Yousuf, 
}
$^{\textcolor{blue}{3}}$\textcolor{blue}{Rajib 
Rana}

$^{\textcolor{blue}{2}}$\textcolor{blue}{Institute 
of Information Technology, 
}$^{\textcolor{blue}{1,3}}$\textcolor{blue}{Institute 
for Resilient Regions (Research)}

$^{\textcolor{blue}{2}}$\textcolor{blue}{Jahangirnagar 
University, Dhaka, Bangladesh, 
}$^{\textcolor{blue}{1,3}}$\textcolor{blue}{University 
of Southern Queensland, Australia}

$^{\textcolor{blue}{1}}$\textcolor{blue}{shezan.huq@gmail.com}

$^{\textcolor{blue}{2}}$\textcolor{blue}{yousuf@juniv.edu}

$^{\textcolor{blue}{3}}$\textcolor{blue}{rajib.rana@gmail.com}
\end{center}

\vspace{4\baselineskip}
\begin{center}
{\large \textbf{Abstract}}
\end{center}

\noindent Image denoising is always a challenging task in the field of computer vision 
and image processing. In this paper we have proposed an encoder-decoder 
model with direct attention, which is capable of denoising and reconstruct 
highly corrupted images. Our model is consisted of an encoder and a decoder, 
where encoder is a convolutional neural network and decoder is a multilayer 
Long Short-Term memory network. In the proposed model, the encoder reads an 
image and catches the abstraction of that image in a vector, where decoder 
takes that vector as well as the corrupted image to reconstruct a clean 
image. We have trained our model on MNIST handwritten digit database after 
making lower half of every image as black as well as adding noise top of 
that. After a massive destruction of the images where it is hard for a human 
to understand the content of those images, our model can retrieve that image 
with minimal error. Our proposed model has been compared with convolutional 
encoder-decoder, where our model has performed better at generating missing 
part of the images than convolutional auto encoder. 

\vspace{1\baselineskip}

\newpage
\section{Introduction}
\noindent Image denoising is a well-studied problem in the field of computer vision and image processing where the task is to take noised images and restore them to the original images. Due to the power of GPU computation, now a day's convolutional neural network is performing very well at image denoising and recognition task \cite{17, 7, 5}. In these models the noised images are filtered by deep convolutional encoder decoder with convolution and deconvolution steps. All of these models are concerned with only the image denoising part, but when a part of the image is missing these models are not very good at generating missing parts as convolutional neural network is not so good at sequence processing. To generate a missing part of the image a model is needed which can understand and predict the upcoming part of the image through reading the current portion of the image. So this model will need memory to capture the sequence and Long Short Term Memory (LSTM) \cite{14} is a well performed recurrent neural network which is good at sequence learning. LSTM also can solve general sequence to sequence problems \cite{15}, which helps to improve machine translation \cite{2} and speech recognition task \cite{3,4}. In the field of natural language processing sequence to sequence mapping with LSTM encoder and decoder performs excellently. Not only in this field but also in the field of machine vision it is performing so well, like in the image description job \cite{8,9}.

\noindent Being inspired by these works we have come up with a CNN encoder and LSTM decoder technique with direct attention. In our model a convolutional neural network reads an image and obtain a fixed size vector representation of the image. Another multilayer LSTM takes that vector and the corrupted image row by row along with the last output from previous timestamp, then it produces the desired image. In our model decoder has the access to the raw image so that if encoder misses any information to capture, it can get it from raw distorted image. The main advantage of our model is that it can clean image as well as can produce lost information. We have trained our model on MNIST handwritten database \cite{19}. For training purpose, we have made lower half of every image as black and then we have added noise on that image. So the image becomes highly destructed and even for a human it is tough to understand what the digit was in that image. We trained our model on these highly distorted images and it was able to remove noises along with the retrieval of the lost information, where convolutional encoder-decoder was able to remove noise but it was unable to find the lost information accurately. As Convolutional neural network is good at cleaning image and LSTM is good at sequence generation, we have combined the power of them to remove the noise from an image and generate missing part of that image.

\section{Related Work}

\begin{figure}
  \includegraphics[width=\linewidth]{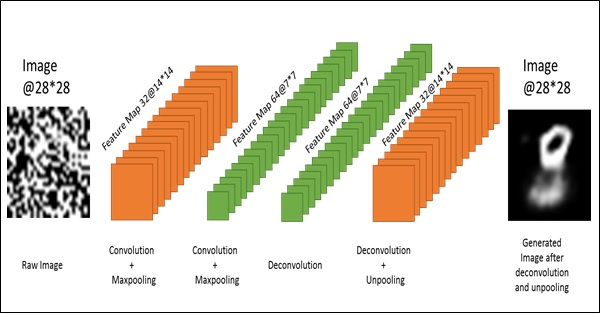}
  \caption{This is an example of typical convolutional encoder decoder.}
  \label{fig:typical_convnet}
\end{figure}

\noindent For image denoising task there has been lot of works till today. But due to the success of deep learning, many deep learning models outperformed the old models. For image denoising task deep convolutional encoder-decoder performs so well \cite{17, 7, 5}. Sample convolutional encoder and decoder is shown in the figure \ref{fig:typical_convnet}.

\noindent These models use a convolutional neural network as encoder which encode the image and symmetric deconvolutional layers as decoder and construct the clean images. For image denoising task these model has outperformed many old models \cite{17}. There is no headache for preprocessing of the image as it is an end to end learning system. It is not common to use convolutional and LSTM together to do the image denoising task. Conv-net and LSTM encoder decoder is a well performed model for dense captioning and image description jobs \cite{8,9}.

\section{Background knowledge}

\noindent For sequence processing Recurrent Neural Network (RNN) \cite{12,1} is a general form of feed forward neural network.  A RNN outputs sequence (y${}_{1}$, y${}_{2}$, {\dots}{\dots}., y${}_{T}$) from the inputs (x${}_{1}$, x${}_{2}$, {\dots}{\dots}., x${}_{T}$) through the iteration over the equation \ref{eq1} and \ref{eq2}

\begin{equation} \label{eq1}
h_t=\sigma \left(w^{h_{x}}x_t+w^{h_{h}}h_{t-1}+b\right)
\end{equation}

\begin{equation} \label{eq2}
y_t={f(w}^{h_{x}}x_t+b)
\end{equation}

\noindent Where,\\
\indent x${}_{t}$   = input at time t \\
\indent h${}_{t-1}$ = input at time t\\
\indent h${}_{t}$   = hidden state at time t \\
\indent y${}_{t}$   = output of time t \\ 
\indent w           = weights \\ 
\indent b           = bias \\
\indent $f$         = any activation function \\ 
\indent \textsigma  = sigmoid activation function \\ 

\begin{figure}[H]
  \includegraphics[width=\linewidth]{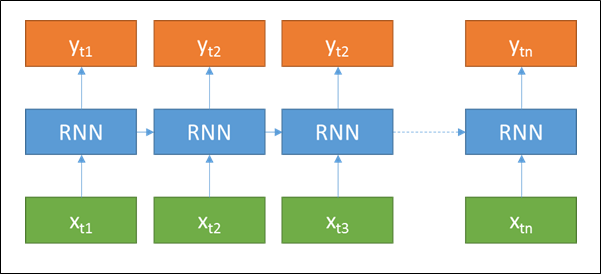}
  \caption{This figure shows the mechanism of a typical Recurrent Neural Network.}
  \label{fig:typical_rnn}
\end{figure}

\noindent In figure \ref{fig:typical_rnn} we can see the the structure of the RNN cells stacked together to process sequence, where at every timestamp there is an output. Due to the vanishing gradient problem it is hard to train recurrent neural network over longer sequence with these settings \cite{13,18}. The Long Short-Term Memory is known to solve these long term dependencies as it has memory state which can carry information longer with the help of input gate, output gate and forget gate \cite{14}. 

\section{Model}
\subsection {Overview of CNN-LSTM encoder decoder with direct attention}
Our model is consisted of an encoder and a decoder. Encoder takes the corrupted image and decoder outputs the cleaned and reconstructed image. So the simplified version of the total system can be expressed through equation \ref{eq3} and \ref{eq4}.

\begin{equation} \label{eq3}
v=f_{encoder}\left({image}_{corrupted}\right)
\end{equation}

\begin{equation} \label{eq4}
{image}_{cleaned}=f_{decoder}\left(v+{image}_{corrupted}\ \right)
\end{equation}

\noindent So the intuition behind this model is, the encoder reads the corrupted image and creates a thought vector where it is supposed to have the full information of the image in the vector. Here encoder has the full privilege to encode anything which will help the model to reduce the loss, so it can encode a clean version of the corrupted image in the vector if it wants.

\begin{figure}[H]
  \includegraphics[width=\linewidth]{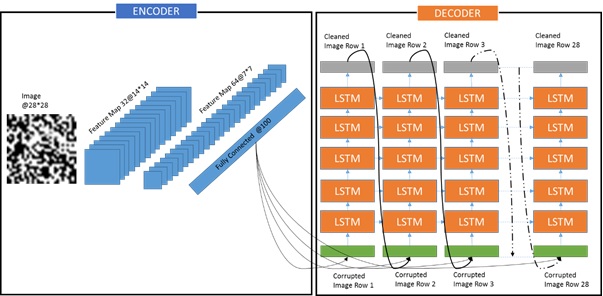}
  \caption{Overall architecture of CNN-LSTM encoder decoder with direct attention.}
  \label{fig:our_model}
\end{figure}

\noindent Then decoder takes that vector and try to produce a clean image where it has the access to that corresponding row of the corrupted image. So overall, when decoder reads an image it has the information about the current row of that image, thought vector and what was the last output by the decoder, so when some part of an image is missing it can reproduce that by its memory like, if lower half of a digit “2” image is missing it can draw that part as it is known to the shape of’ “2”. Figure \ref{fig:our_model} shows the architecture of the whole model. The encoder part of the figure shows the convolutional neural network architecture of the model of following 32, Maxpooling, 64, Maxpooling, fully connected layers and the decoder part shows the 5 layers LSTM which produces the final image row by row.

\subsection{Detail of the Architecture}
\subsubsection{Encoder}

The Encoder of the model is consist of a convolutional neural network. The output from the encoder is a fully connected layer. If the image is represented as x, then filter k with size n${}_{k}$, convolve over the different channel of the image with stride of size s${}_{k}$. The weights of the filter k is shared spatially and they are different for every channel of the feature map. There are subsampling/pooling layers to shrink the width and height of the feature map with filter size p and stride size s${}_{p.  }$The equation for convolutional layer \cite{6} is

\begin{equation} \label{eq5}
y_i{=f}_a(\sum_i{{\mathrm{k}}_{\mathrm{ij}}{\mathrm{*}\mathrm{x}}_{\mathrm{i}})}
\end{equation}

\noindent Where, x${}_{i}$ is the i${}^{th}$ channel of input, k${}_{ij}$ is the convolution kernel, y${}_{j}$ is the hidden layer, $f_a$ is activation function.

\noindent The equation \ref{eq6} for pooling layer is

\begin{equation} \label{eq6}
y_{ijk}=\smash{\displaystyle\max_{p,q}} \quad x_{i,j+p,k+q}
\end{equation}

\noindent Here, x${}_{i,j,k}$ is value of the i${}^{th}$ feature map at position $j$, $k$ ; $p$ is vertical index in local neighborhood, $q$ is horizontal index in local neighborhood, y${}_{ijk}$ is pooled and sub sampled layer.

\noindent The fully connected layer is achieved through a dot product between the final layer $y$  and weight matrix $\ W^v$ and then adding bias vector $b$. Then the output is passed through any activation function $f_a$.

\begin{equation} \label{eq7}
v=\ {f_a(W}^vy+b)
\end{equation}

\noindent Final fully connected layer is the though vector $v$ which is passed to the decoder to draw the final image.

\subsubsection{Decoder}
The Decoder of our model is consist of multilayer LSTM.  The LSTM is consist of Input gate, Output gate, Forget gate and New Memory cell. The Input gate, Output gate, Forget gate and New Memory cell/Current State is generated from the raw input, precious hidden state, thought vector from encoder and previous output from decoder. 

\noindent The input gate decide the passing of new memory where the forget gate decide whether to keep the previous memory or not. The output gate decide the outflow of the current hidden state. 

\noindent In our decoder the initial hidden state and memory state is zero. Initial bottom LSTM cells of the multilayer LSTM, takes the thought vector produced by encoder, current row of the corrupted image and the output image row from its previous units. 

\noindent $W$ denotes as the Weight and $b$ denotes as the Bias. The equations \ref{8} to \ref{13} for the LSTM cell of the decoder is stated below (Biases are not shown in the equations)

\begin{equation} \label{eq8}
Input \ gate,\ i^{t}=\sigma (x_{t}W^{x_{i}}+h_{t-1}W^{h_{i}}+vW^{v_{i}}+y_{t-1}W^{y_{i}})
\end{equation}

\begin{equation} \label{eq9}
Forget \ gate,\ f^{t}=\sigma (x_{t}W^{x_{f}}+h_{t-1}W^{h_{f}}+vW^{v_{f}}+y_{t-1}W^{y_{f}})
\end{equation}

\begin{equation} \label{eq10}
Output \ gate,\ o^{t}=\sigma (x_{t}W^{x_{o}}+h_{t-1}W^{h_{o}}+vW^{v_{o}}+y_{t-1}W^{y_{o}})
\end{equation}

\begin{equation} \label{eq11}
New \ state,\ \widetilde{C_{t}}=\tanh (x_{t}W^{x_{c}}+h_{t-1}W^{h_{c}}+vW^{v_{c}}+y_{t-1}W^{yc})
\end{equation}

\begin{equation} \label{eq12}
Final \ state,\ C_{t} = C_ {t-1}\circ f^t + \widetilde{C_{t}} \circ i^t
\end{equation}

\begin{equation} \label{eq13}
Final \ hidden \ state,\ h_t  = \tanh (C_{t}) \circ i^{t}
\end{equation}

\noindent Here, $x_t$ is the current row of the corrupted image, $h_{t-1}$ is the previous hidden state, v is the thought vector and $y_{t-1}$ is the last output of the decoder. The output for every row is derived through below equation

\begin{equation} \label{eq14}
Output,\ y_{t} = \sigma(h_{t}W^{y_{i}} + b)
\end{equation}

\subsubsection{Loss function}
The loss function is the mean squared loss. The loss is given by,

\begin{equation} \label{eq15}
Loss \ =  \sqrt {\dfrac {1} {n}\sum _{i}\left( \hat{y}-y\right) }
\end{equation}

Where, $\hat{y}$  is the predicted output and y is the original output. 

\newpage
\subsubsection{Activation Functions}
For the convolutional layer of the decoder, activation function Relu \cite{11} is used.

\begin{equation} \label{eq16}
Relu,\ f(x) = \max(0, x)
\end{equation}

\begin{figure}[H]
  \includegraphics[width=\linewidth]{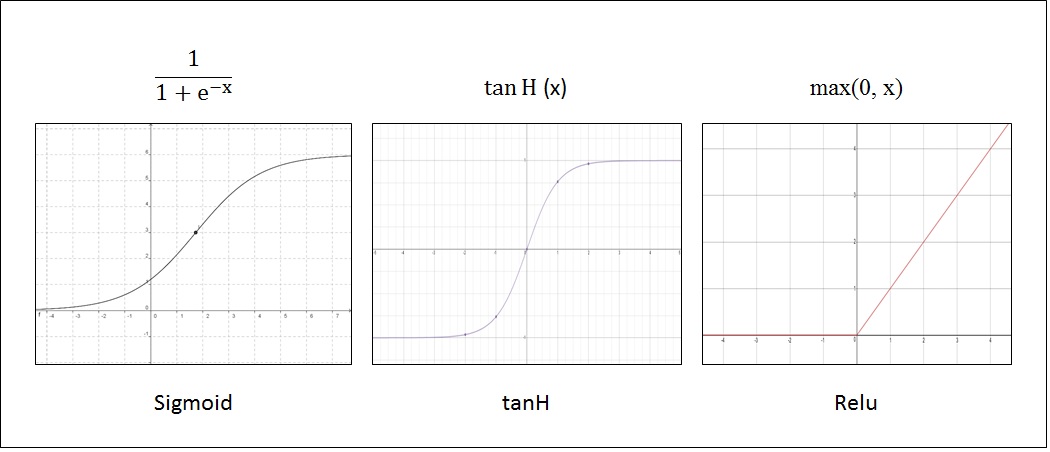}
  \caption{This figure shows three different activation function’s graphs and its equations. From left sigmoid, $\tanh$ and Relu is shown in this figure.}
  \label{fig:activation_function}
\end{figure}

\noindent For the final fully connected layer of the decoder $\tanh$ activation function is used. 
In decoder for the final output layer Sigmoid activation function is used to make the output very near to the original image as the image was normalized thorough making the pixel values between 0 and 1. Figure \ref{fig:activation_function} shows different activation functions.  

\subsubsection{Regularizations}

For the regularization purpose, L2 loss is used. The L2 loss is shown in equation \ref{eq17}.

\begin{equation} \label{eq17}
L2=\lambda \sum^k_{i=1}{{\omega }^2_i}
\end{equation}

\noindent Where, $\omega $ represents the weights and $\lambda $ is a hyper parameter. How much L2 loss is added to the final loss of the mode is determined by$\lambda $.

\subsubsection{Optimization}
\noindent The weights of the model is trained with both Adadelta optimization cite{20} and Adam optimizer \cite{21}. Finally it is observed that Adam performed well over Adadelta.

\section{Training Details}

\subsection{Dataset Preparation}
\begin{figure}[H]
  \includegraphics[width=\linewidth]{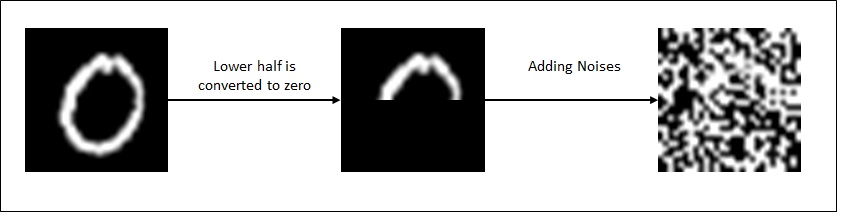}
  \caption{This figure show the training data generation process. First the lower 14 rows were made blank then top of the whole image the noises were added.}
  \label{fig:trainning_process}
\end{figure}

\noindent For the training purpose the MNIST handwritten digit dataset was used. We have made some the lower half of the image as blank/ black. After that we have added Salt-and-pepper noise or white noise top of that image, which made the image highly distorted. Figure \ref{fig:trainning_process} demonstrate the training data generation process. We have divided the data into train and test where train contains 75 \% of the data and 25 \% of the data is belongs test set.

\subsection{Configuration}

For the proposed CNN-LSTM model, after trying different combinations the selected Encoder’s architecture is shown in Table \ref{t1}

\begin{table}[H]
\centering
\caption{Selected Encoder architecture}
\label{t1}
\begin{tabular}{|c|c|c|c|}
\hline
Layer Name      & \begin{tabular}[c]{@{}c@{}}Input Shape\\ (depth@hight*weight)\end{tabular} & \begin{tabular}[c]{@{}c@{}}Filter Size\\ (depth@hight*weight)\end{tabular} & \begin{tabular}[c]{@{}c@{}}Result\\ (depth@hight*weight)\end{tabular} \\ \hline
Conv 1          & 1@28*28                                                                    & 32@5*5                                                                     & 32@28*28                                                              \\ \hline
MaxPool 1       & 32@28*28                                                                   & 2*2                                                                        & 32@14*14                                                              \\ \hline
Conv 2          & 32@14*14                                                                   & 64@5*5                                                                     & 64@14*14                                                              \\ \hline
MaxPool 2       & 64@14*14                                                                   & 2*2                                                                        & 64@7*7                                                                \\ \hline
Fully Connected & 64*7*7                                                                     & 64*7*7*100                                                                 & 100                                                                   \\ \hline
\end{tabular}	
\end{table}

\noindent The decoder was five-layered LSTM. We have also trained a convolutional encoder-decoder (CNN) of 32, 64 convolutional layer and symmetric deconvolutional layer for comparison. Figure \ref{fig:typical_convnet} shows the architecture of this model.

\subsection{Other Details}

We have written the code with the help of deep learning framework “Tensorflow” provided by Google \cite{10}.  The images were edited with python programming language. We have run both of the model on the same dataset and the iteration was 500k with batch size 100. Dropout \cite{16} of 25\% was used to deter the model from over fitting.

\section{Results}
Proposed CNN-LSTM model and CNN-CNN model was trained on Nvidia GTX 1080 Gpu, for 500k iteration of batch size 100. Required time for both models is shown in Table \ref{t2}.

\begin{table}[H]
\centering
\caption{Required time for both models}
\label{t2}
\begin{tabular}{|c|c|}
\hline
Model Name & Required Time                     \\ \hline
CNN-CNN    & 4 Hours 19 Minutes and 52 seconds \\ \hline
CNN-LSTM   & 6 Hours 58 Minutes and 24 seconds \\ \hline
\end{tabular}
\end{table}

\begin{figure}[H]
  \includegraphics[width=\linewidth]{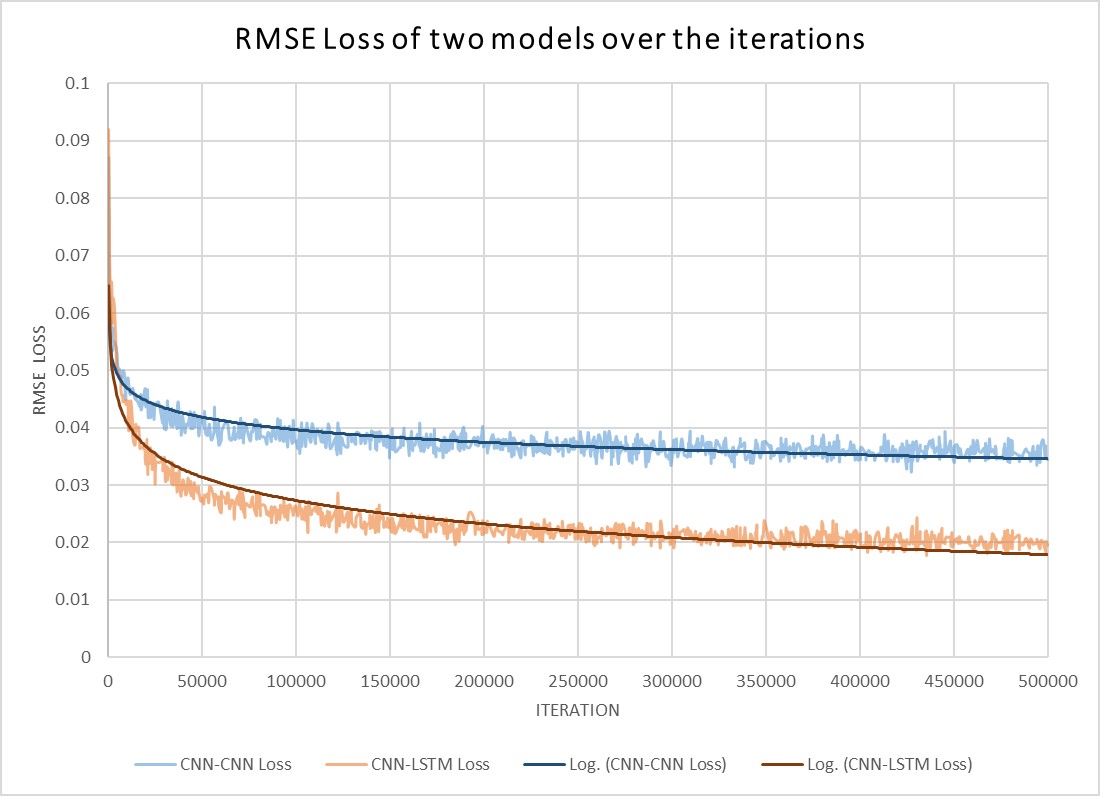}
  \caption{Movement of loss for both CNN-CNN and CNN-LSTM models. A log trend line is fitted over the loss of the models.}
  \label{fig:loss}
\end{figure}

\noindent Loss for both model is depicted in Figure \ref{fig:loss}. The graph shows the movement of loss for both CNN-CNN and CNN-LSTM models. It is observed from the graph that both losses have a jumping behavior throughout the training period and it happened as it was stochastic training process. We fitted a log trend line to see the smooth movement of the loss. The loss of CNN-CNN was unable to improve after certain iteration but for CNN-LSTM, the loss was decreasing over the iterations. From the graph it can be inferred that after 5000 iterations the loss might goes down.

\begin{figure}[H]
  \includegraphics[width=\linewidth]{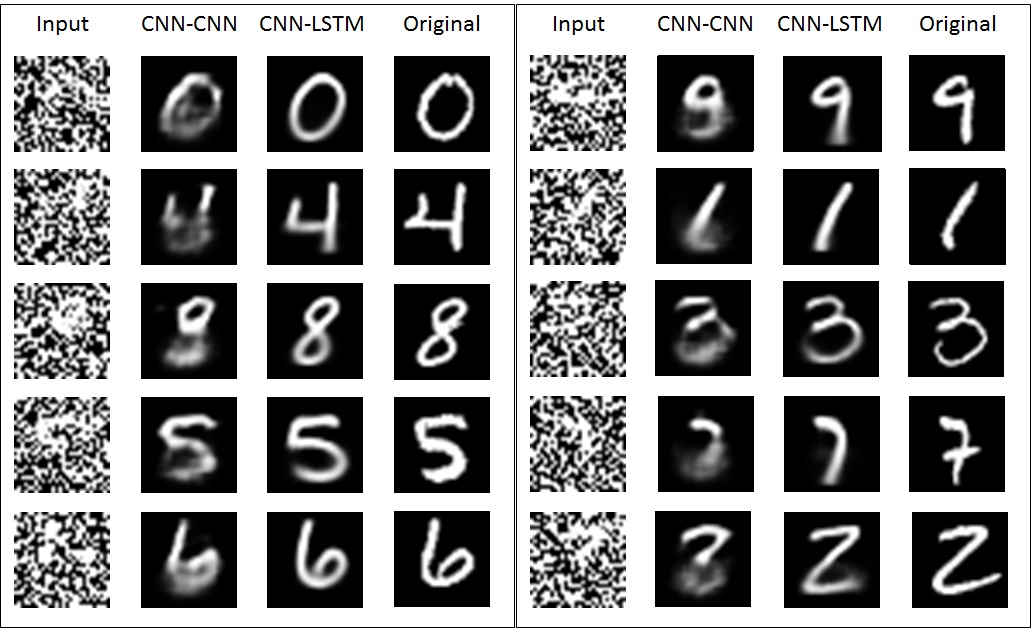}
  \caption{This figure shows the performance of CNN-CNN and CNN-LSTM models. Input column represent the corrupted images, CNN-CNN column shows the CNN-CNN model generated images, CNN-LSTM shows the CNN-LSTM model generated images and Original column represents the original images.}
  \label{fig:result}
\end{figure}

\noindent Our model was capable of removing noise from the image as well as it was capable of retrieving the lost part of the images with minimal error. The difficult task was to draw the lost shape of the digits and CNN-LSTM performed outstandingly and for most of the case generated the lost shape with great perfection where even for a human eye it was too tough to understand the content of the distorted images. Figure \ref{fig:result} shows the performance of the proposed model and comparison with the CNN-CNN model.   
\noindent There is an amazing fact to notice that our CNN-LSTM model with direct attention, produced cleaned and fined image. The edge of the produced images are not as rough as the original. Our model produced smoothed images. This amazing result happed due to the RMSE loss. Due to RMSE loss the model tried to predict pixel values which is not far from original value. The width of the stroke of the images are different and it has not pattern to learn that’s why the model learned the smoothed shape so that it minimize the loss for all types of strokes.  

\noindent Convolutional encoder-decoder performs very well at image denoising task but for the generation of the missing part it was not so good. So our model has outperformed the convolutional encoder and decoder in the part of generating the missing part of the images. Convolutional neural network is very good at denoising and that’s why we have used CNN encoder and LSTM decoder as this is very good at sequence generation. 

\section{Conclusion}
Image denoising is an active field of research in the area of image processing. If some portion of the image is missing then almost all algorithms try to produce that missing part from surrounding pixels. If major portion of an image is missing then the missing part cannot be produced through observing the neighborhood pixels. To produce a missing part from an image the knowledge about that object is necessary. If half of a digit is missing then any algorithm needs to understand what that digit looks like, then it can produce the missing part keeping the relevance with the current portion of the image. The algorithm need to learn the generic shape of a digit to do this job. CNN-CNN encoder decoder outperformed many models in terms of image denoising but it is not good at producing missing part of an image as it has no memory. Among deep learning models Recurrent Neural Networks are good at sequence processing as they have memory module. To combine the power of CNN and RNN model we proposed CNN-LSTM Encoder Decoder with Direct Attention model, which is very good at denoising images as well producing lost part of the image. The model was trained on MNIST handwritten digit dataset after massive distortion. The goal of our model is to learn to remove noise and the shape of digit and our model has done that with minimal errors which can be removed through long training time.

\noindent have achieved an outstanding result from our model we are researching on Generative Adversarial Networks (GAN) \cite{22}. Due to exemplary performance of GAN \cite{23,24}, we are trying to use GAN instead of RMSE loss layer, which might be helpful to converge our model earlier with better result.

\end{document}